\documentclass[a4paper, 11pt]{article}

\usepackage{graphicx}
\usepackage[margin=1.3in,footskip=0.25in]{geometry}
\begin{document}

%%%%%%%%%%%%%%%%%%%%%%%%%%%%%%%%%%%%%%%%%%%%%%%%%%%%%%%%%%%%%%%%%%%%%%%

%Title
\title{Noise impact on recurrent neural network with linear activation function}

%Authors for header
\author{V.\,M.\,Moskvitin, N.\,Semenova}

\author{V.\,M.\,Moskvitin, N.\,Semenova}

%\keywords{artificial neural networks, recurrent neural network, echo state network, noise, dispersion, statistic, white Gaussian noise}

\maketitle

%Abstract
\begin{abstract}% about 100 words
Over the past few years, artificial neural networks (ANNs) have found their application in solving many problems from pattern recognition to predicting climate phenomena. Despite the existence of high-power computing clusters with the ability to make parallel calculations, neural network modeling on digital equipment is a bottleneck in network scaling and the speed of obtaining and processing information. In recent years, more and more researchers in the field of neural networks are interested in creating hardware implementations where neurons and the connection between them are realized physically. 

The physical implementation of ANN fundamentally changes the features of noise influence. In the case hardware ANNs, there are many internal sources of noise with different properties. The purpose of this paper is to study the peculiarities of internal noise propagation in recurrent ANN on the example of echo state network (ESN), to reveal ways to suppress such noises and to justify the stability of networks to some types of noises. 

In this paper we analyse ESN in presence of uncorrelated additive and multiplicative white Gaussian noise. Here we consider the case when artificial neurons have linear activation function with different slope coefficients. Starting from studying only one noisy neuron we complicate the problem by considering how the input signal and the memory property affect the accumulation of noise in ESN. In addition, we consider the influence of the main types of coupling matrices on the accumulation of noise. So, as such matrices, we take a uniform matrix and a diagonal-like matrices with different coefficients called ``blurring’’ coefficient. 

We have found that the general view of variance and signal-to-noise ratio of ESN output signal is similar to only one neuron. The noise is less accumulated in ESN with diagonal reservoir connection matrix with large ``blurring'' coefficient. Especially it concerns uncorrelated multiplicative noise. 
\end{abstract}

%\newpage

\section{Introduction}
\setcounter{equation}{0} 

Over the past few years, artificial neural networks (ANNs) have been applied in solving many problems \cite{LeCun2015}. Such tasks include image recognition \cite{Maturana2015, Krizhevsky2017}, image classification, improvement of sound recordings, speech recognition \cite{Graves2013}, prediction of climatic phenomena \cite{Kar2009} and many others.

The basic principle of ANN construction is signal propagation between neurons using connections with some coefficients. In this case, the greatest efficiency and speed can be achieved by paralleling calculations on high-performance computing clusters. However, in this case the bottleneck is the speed of memory access and data processing. The maximum performance of calculations can be achieved only in case if ANN is completely hardware-implemented. In this case, the problem of memory access and mathematical operations over a large amount of data disappears, since each neuron corresponds to an analog nonlinear component, and each connection to a physical connection channel. 

In recent years, there has been an exponential increase in work with hardware implementations of ANNs. Currently, the most effective ANNs are based on lasers \cite{Brunner2013}, memristors \cite{Tuma2016}, and spin-torque oscillators \cite{Torrejon2017}. Connection between neurons in optical ANN implementations is based on the principles of holography \cite{Psaltis1990}, diffraction \cite{Bueno2018, Lin2018}, integrated networks of Mach-Zender modulators \cite{Shen2017}, wavelength division multiplexing \cite{Tait2017}, and 3D printed optical interconnects \cite{Moughames2020,Dinc2020,Moughames2020a}. Recently, the so-called photonic neural networks are gaining popularity \cite{MourgiasAlexandris2022, Wang2022}.

The physical implementation of ANN fundamentally changes the features of noise influence. In the case of digital computer implementation of ANN, noise can enter the system exclusively with the input signal, whereas in analog ANN there are many internal sources of noise with different properties. The purpose of this paper is to study the peculiarities of internal noise propagation in recurrent ANN, to reveal ways to suppress such noises and to justify the stability of networks to some types of noises. 

In our previous studies we were focused on the effects of additive and multiplicative, correlated and uncorrelated noise on deep neural networks \cite{Semenova2019,Semenova2022}. Several models of varying complexity were considered. General features depending on the nonlinear activation function and the depth of the ANN were shown for simplified symmetric ANNs with global uniform connectivity between layers.  All the findings and results were then validated for three trained deep ANNs used for number recognition, classification of clothing images, and chaotic realization predictions. Using the analytical methods described in Ref.~\cite{Semenova2022}, several noise reduction strategies were proposed in our next study \cite{Semenova2022a}.

In this work, we make a task of noise study more complicated by considering time-dependency. In contrast to previously considered deep neural networks, here we are focused on recurrent neural network on the example of echo state neural network (ESN). This network consists of three main parts: 1-- input layer receiving the input signal and transmitting it to the next layer; 2 -- one layer called reservoir which state depends on both input signal at current moment and previous states of reservoir at previous times; 3 -- output layer making the final output signal. Such networks are often used to work with signals that are highly dependent on time. For example, prediction of chaotic temporal realizations, speech recognition, etc.

\section{System under study}

\subsection{Noise types}\label{sec:noise_types}
In this paper we consider only white Gaussian noise with zero mean and some constant dispersion $D$. The noise values will be different for each neuron each time, so it is uncorrelated in time and in network. Mathematically speaking it is introduced into each artificial neuron according to the noise operator $\mathbf{\hat N}$ as
\begin{equation}\label{eq:neuron_noise}
    y_i(t)=\mathbf{\hat N}x_i(t)=x_i(t)\cdot(1+\xi_M(t,i)+\xi_A(t,i),
\end{equation}
where $x_i$ and $y_i$ are noise-free and noisy outputs of the $i$th artificial neuron, respectively. $\xi$ are the sources of white Gaussian noise with zero mean. The indices 'A' and 'M' points out the noise types, namely additive (A) and multiplicative (M) noise with noise dispersions $D_A$ and $D_M$. As can be seen from (\ref{eq:neuron_noise}), the additive noise is added to the noise-free output, while the multiplicative noise is multiplied on it. The part $(1+\dots)$ is needed to keep the useful signal. The notation of noise operator $\mathbf{\hat N}$ will be used further to indicate which outputs of neurons become noisy. 

The noise dispersions will be fixed throughout the paper as $D_A=D_M=10^{-2}$. This order of values corresponds to what we have previously obtained in an RNN realized in optical experiment \cite{Brunner2013, Semenova2019}.

\subsection{Recurrent neural network}\label{sec:RNN_definition}
\begin{figure}[!ht]
\centering\includegraphics[width=\linewidth]{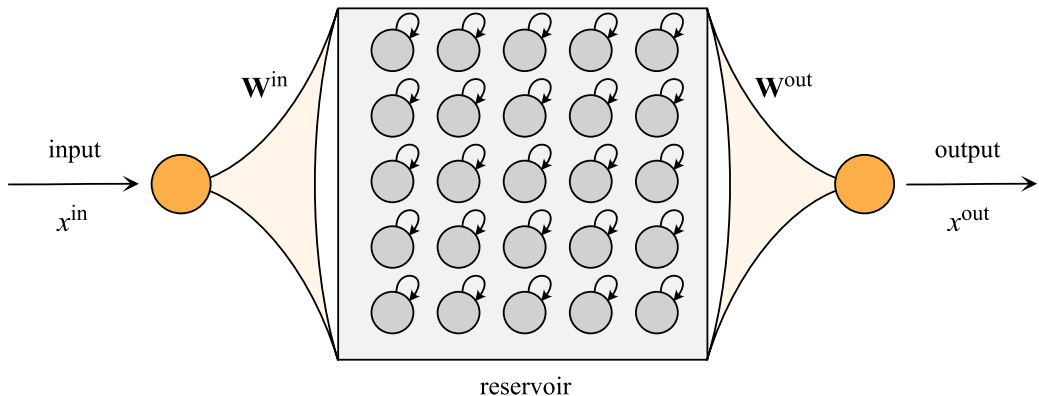} 
\caption{Schematic representation of considered recurrent neural network.} \label{fig:scheme}
\end{figure}

There are many different types of neural networks. Their topology and type of neurons strongly depend on a signal type and features of the tasks being solved. If the network is trained to work with signals changing in time, such as speech recognition, prediction of chaotic phenomena etc., then the neural network must have a property of memory. Here come to the aid of recurrent neural networks (RNNs). In RNNs, the part of neurons have the memory about their previous states. In this paper we consider an echo-state network (ESN) schematically shown in Fig.~\ref{fig:scheme} as an examples of RNN. This network contains input and output neurons (orange) and a hidden layer with multiple neurons called reservoir (gray). The connectivity and weights of neurons inside the reservoir $\mathbf{W}^\mathrm{res}$ are usually fixed and randomly assigned. The output connection matrix $\mathbf{W}^\mathrm{out}$ is varied during training process to make the network producing correct responses to certain input signals. 

In this paper we are mainly interested in the impact of noise on reservoir part of the network, and therefore, only gray neurons have a noise influence.

In accordance with notations in Fig.~\ref{fig:scheme}, the input signal $x^\mathrm{in}$ passes thought the input neuron and comes to reservoir neurons coupled via the input matrix $\mathbf{W}^\mathrm{in}$ of size $(1\times N)$, where $N$ is the number of neurons inside reservoir. It is fixed throughout the paper as $N=100$. The reservoir neurons have the connection to their previous states via connection matrix $\mathbf{W}^\mathrm{res}$ of size $(N\times N)$. Thus, the equation of reservoir neurons is
\begin{equation}\label{eq:neuron_reservoir}
    \mathbf{x}^\mathrm{res}_t=f\big( \beta x^\mathrm{in}_t\cdot \mathbf{W}^\mathrm{in} + \gamma \mathbf{y}^\mathrm{res}_{t-1}\cdot \mathbf{W}^\mathrm{res} \big); \ \ \ \ \ \ \ \mathbf{y}^\mathrm{res}_t=\mathbf{\hat N}\mathbf{x}^\mathrm{res}_t,
\end{equation}
where $f(\cdot)$ is the activation function of reservoir neurons. The type of activation function often depends on the current task. In this paper we are mainly focused on linear activation function $f(x)=\alpha x$, since linear and partly linear functions are often used in RNNs. The nonlinear activation function can lead to completely different dynamics and noise accumulation, and it will be therefore a subject of another study.

The index $t$ corresponds to the current time moment, while $(t-1)$ in the term $\mathbf{y}^\mathrm{res}$ indicates that the outputs of reservoir neurons are taken from the previous time moment. The bold font used for $\mathbf{y}^\mathrm{res}_t$ and $\mathbf{y}^\mathrm{res}_t$ indicates that they are row-vectors $(1\times N)$.

Parameters $\beta$ and $\gamma$ control the impact of input signal ($\beta$) and memory ($\gamma$). In order to keep the same range of the output signals, the condition $\beta+\gamma=1$ is imposed on them.

The output of the network comes from the output neuron connected with reservoir via connection matrix $\mathbf{W}^\mathrm{out}$ of size $(N\times 1)$:
\begin{equation}\label{eq:neuron_out}
    x^\mathrm{out} = \mathbf{y}^\mathrm{res}_t\cdot \mathbf{W}^\mathrm{out}.
\end{equation}
In order to see the pure impact and statistics of noise, the output connection matrix $\mathbf{W}^\mathrm{out}$ is fixed and uniform with elements $1/N$. The input connection matrix $\mathbf{W}^\mathrm{in}$ is responsible for sending an input signal to the reservoir. That is why its values are set to be unity.

\section{One noisy neuron}\label{sec:one_neuron}
As a first step let us consider how noise impacts on one isolated neuron with linear activation function. The neuron receives the input signal $x^\mathrm{in}_t$ and produces the noise-free signal $x^\mathrm{out}_t=f(x^\mathrm{in}_t)$ becoming $y^\mathrm{out}_t=\mathbf{\hat N}x^\mathrm{out}_t$ after the noise influence. The input signal contains $T=200$ random values from $-1$ to $1$. 

\begin{figure}[!ht]
\centering\includegraphics[width=1\linewidth]{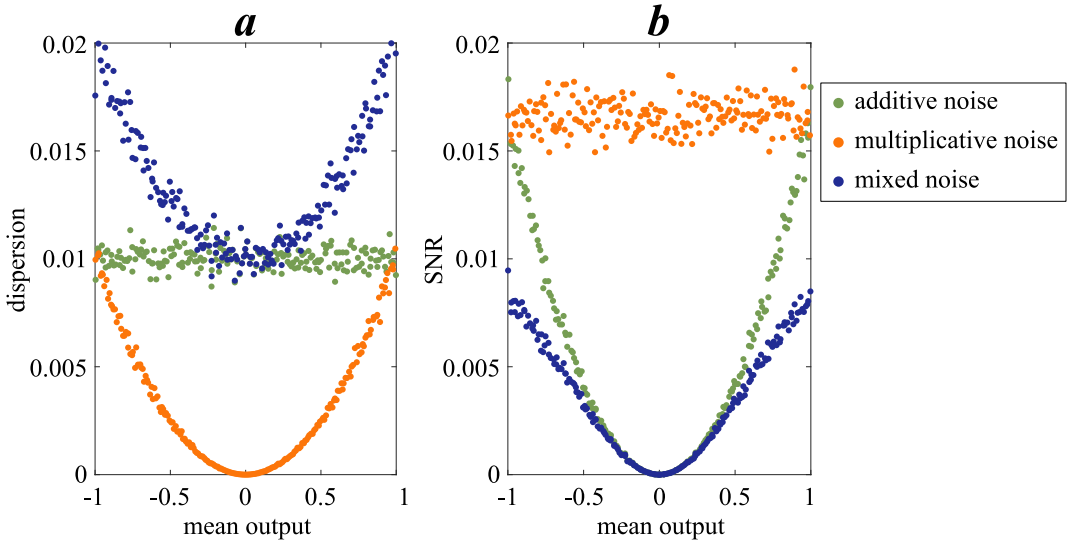} 
\caption{Dispersion $\sigma^2$ (panel a) and SNR (panel b) calculated for the output of one noisy neuron with only additive noise (green points), only multiplicative noise (orange) and mixed noise (blue) depending on corresponding mean output value. The dispersion of noise sources are $D_A=D_M=10^{-2}$. The neuron has linear activation function with $\alpha=1$.} \label{fig:one_neuron}
\end{figure}

Figure \ref{fig:one_neuron} shows the impact of additive (green), multiplicative (orange) and mixed (blue) noise.

In order to see the noise impact we will use the next two characteristics. The first one is a dispersion showing the dispersion of the output signal. It is calculated in the next way. Each input signal $x^\mathrm{in}_t$ is repeated $K=1000$ times to get the statistics of output values. Then the obtained sequence of noisy output values is averaged to get the corresponding mean value $\mu(y^\mathrm{out}_t)$ and dispersion $\sigma^2(y^\mathrm{out}_t)$. The dispersion for different noise types is given in Fig.~\ref{fig:one_neuron}a.

The second characteristics is a signal-to-noise ratio (SNR) showing the relation between the mean output value and its variance or dispersion. In our previous papers \cite{Semenova2019, Semenova2022} we calculated the characteristics similar to SNR, as working for only positive input and output values. Now we will consider both positive and negative values and use therefore more common form of SNR calculated as $\mathrm{SNR}[y^\mathrm{out}]=\mu[y^\mathrm{out}]/\sigma^2[y^\mathrm{out}]$ (see Ref. \cite{snr}). The SNR for different noise types is given in Fig.~\ref{fig:one_neuron}b.

As can be seen from Fig.~\ref{fig:one_neuron}a, the additive noise (green points) lead to the constant level of dispersion which does not depend on the input and output signal. The corresponding dispersion can be found as the variance of a random signal:
\begin{equation}\label{eq:one_neuron_add}
    \sigma^2[y^\mathrm{out}_t]=\mathrm{Var}[y^\mathrm{out}_t]=\mathrm{Var}[f(x^\mathrm{in}_t)+\xi_A]=\mathrm{Var}[\xi_A]=D_A.
\end{equation}
In the case of multiplicative noise, the variance becomes completely different (orange). There is a quadratic relationship between mean output value and its variance. In terms of expectation value and variance, the variance of the output signal of one neuron with multiplicative noise can be found as follows
\begin{equation}
    \mathrm{Var}[y^\mathrm{out}_t]=\mathrm{Var}\big[f(x^\mathrm{in}_t)\cdot(1+\xi_M)\big]=\big(\mathrm{E}[f(x^\mathrm{in}_t)]\big)^2\cdot\mathrm{Var}[\xi_A]=D_A\cdot \big(\mathrm{E}[y^\mathrm{out}_t]\big)^2,
\end{equation}
where $\mathrm{E}[\cdot]$ is the expectation value. The expectation value of the output signal is $\mathrm{E}[y^\mathrm{out}_t]=\mathrm{E}[x^\mathrm{out}_t]=\mathrm{E}[f(x^\mathrm{in}_t)]$.

In the case of linear activation function the dispersion strongly depends on the parameter $\alpha$:
\begin{equation}\label{eq:one_neuron_mul}
    \sigma^2[y^\mathrm{out}_t]=\mathrm{Var}[y^\mathrm{out}_t]=\mathrm{Var}\big[f(x^\mathrm{in}_t)\cdot(1+\xi_M)\big]=D_A(\alpha x^\mathrm{in}_t)^2.
\end{equation}

Mixed noise (blue points in Fig.~\ref{fig:one_neuron}) combines features of both additive and multiplicative noise. Thus, the dispersion (panel a) is the sum of additive and multiplicative variances. For this reason, the SNR in the case of mixed noise with $D_A=D_M$ is reduced twice (panel b).

\begin{figure}[!ht]
\centering\includegraphics[width=1\linewidth]{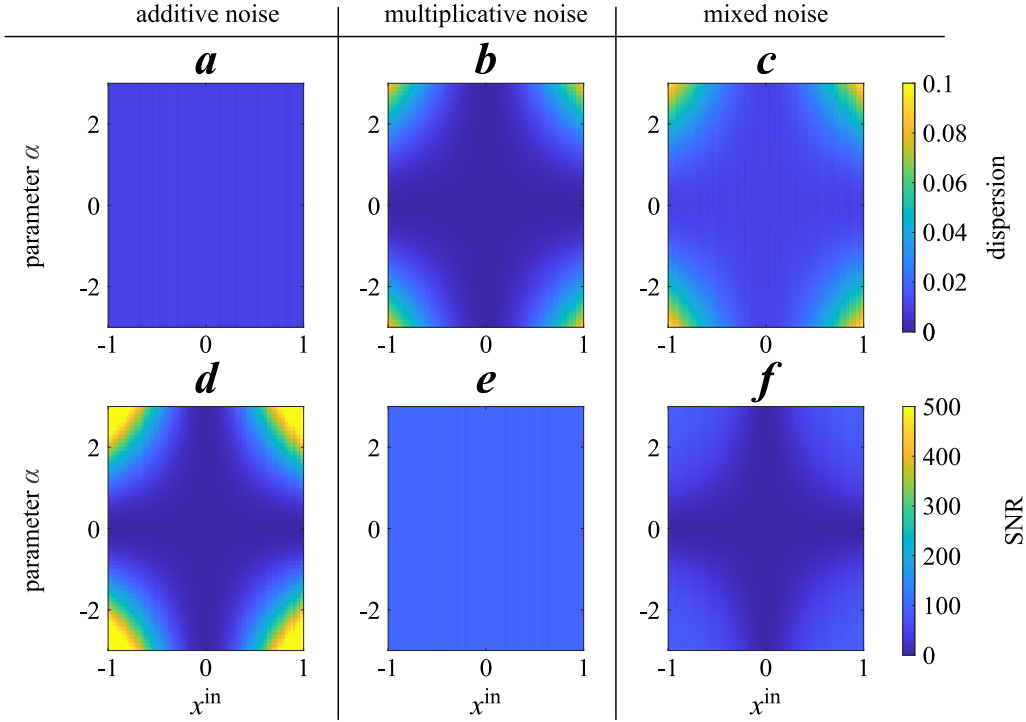} 
\caption{Dispersion (panels a--c) and SNR (panels d--f) calculated for the output of one noisy neuron with only additive noise (a,d), only multiplicative noise (b,e) and mixed noise (c,f) depending on input value $x^\mathrm{in}$ and parameter $\alpha$. The dispersion of noise sources are $D_A=D_M=10^{-2}$.} \label{fig:one_neuron_alpha}
\end{figure}

The impact of $\alpha$-parameter on SNR and dispersion is shown in Fig.~\ref{fig:one_neuron_alpha}. It confirms Eqs. (\ref{eq:one_neuron_add},\ref{eq:one_neuron_mul}). The input signal and $\alpha$ does not change the dispersion in case of additive noise (Fig.~\ref{fig:one_neuron_alpha}a) and SNR in case of multiplicative noise (Fig.~\ref{fig:one_neuron_alpha}e). The parameter $\alpha$ has no impact on SNR with additive noise (Fig.~\ref{fig:one_neuron_alpha}d), while both $\alpha$ and input signal change the dispersion in case of multiplicative noise (Fig.~\ref{fig:one_neuron_alpha}b).

\section{ESN with uniform connection matrix}\label{sec:ESN_uniform}
In this section, we will focus on the interplay of input signal and memory. In order to see pure noise accumulation without the impact of connectivity matrices, we consider the uniform connection matrix in reservoir: $W^\mathrm{res}_{ij}=1/N$.

As a first step, we set $\gamma=0$, when the reservoir has no memory, and the state of reservoir depends only on the input signal. According to Sect.~\ref{sec:RNN_definition}, if $\gamma=0$, then $\beta=1$.

If the property of memory is turned off, then how the input signal depends on time is irrelevant. Therefore, we use the same random input signal from -1 to 1 as in the previous section. 

\begin{figure}[!ht]
\centering\includegraphics[width=1\linewidth]{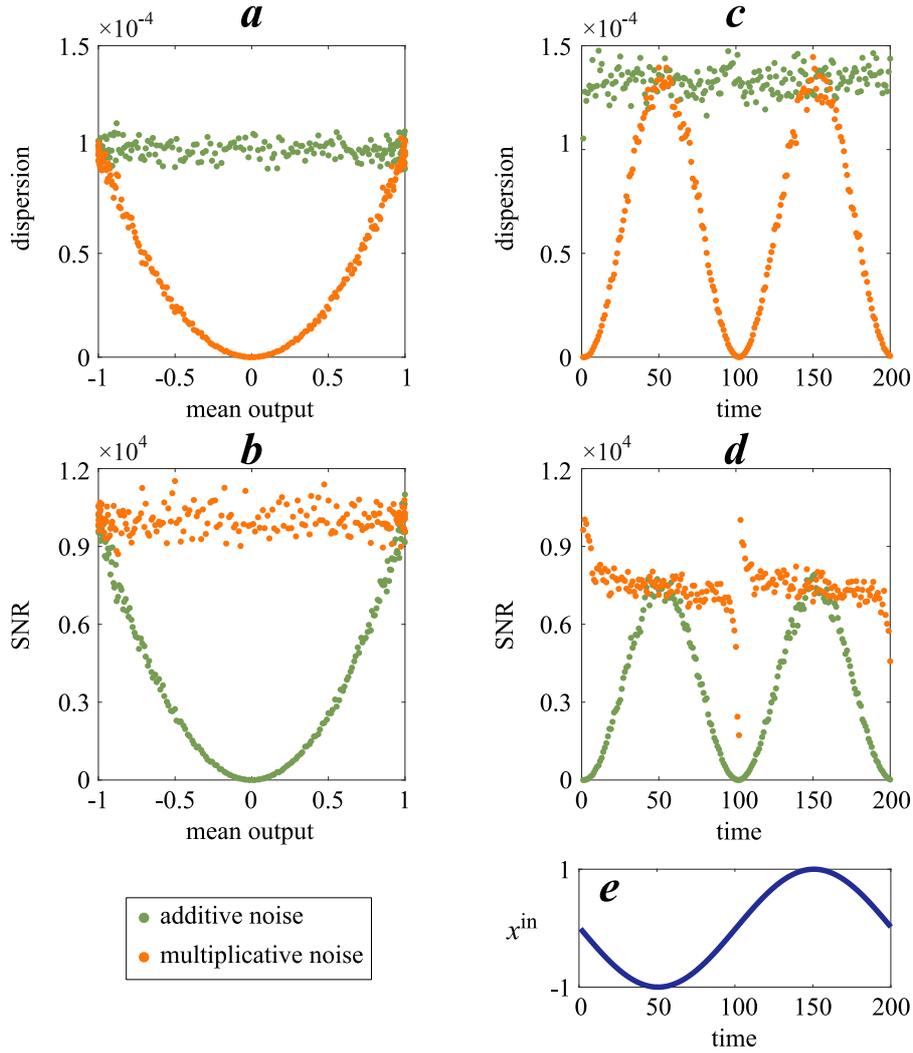} 
\caption{Dispersion (a,b) calculated for the output of ESN with $\gamma=0$ (left panels) and $\gamma=0.5$ (right panels). The input signal (c) was used when $\gamma\neq 0$. Parameters: $\alpha=1$, $\beta=1-\gamma$, $D_A=D_M=10^{-2}$.} \label{fig:ESN_uni}
\end{figure}

Figure \ref{fig:ESN_uni}a shows the dispersions calculated by the output signal of ESN with $\gamma=0$ for additive (green) and multiplicative (orange) noise. The general view of these dependencies is the same as was for one noisy neuron. The difference is the range of these values. Comparing Figs. \ref{fig:one_neuron}a and \ref{fig:ESN_uni}a, the dispersion of ESN's output is reduced by a factor of $100$. Thus, the final output signal becomes less noisy.

This can be explained as follows. We introduce the noise only to neurons of the reservoir. In the case of only additive noise their dispersion can be calculated according Eq.~(\ref{eq:one_neuron_add}). If $\gamma=0$, then the output signal can be calculated as $x^\mathrm{out}_t=\frac{1}{N}\sum\limits^N_{j=1} x^\mathrm{res}_{t,j}$. Then the output dispersion and variance for additive noise is 
\begin{equation}\nonumber
    \mathrm{Var}[x^\mathrm{out}_t] = \Big(\frac{1}{N}\Big)^2 \sum\limits^N_{j=1} \mathrm{Var}[\alpha x^\mathrm{in}_t+\xi_A(j,t)] = \Big(\frac{1}{N}\Big)^2 \sum\limits^N_{j=1} \mathrm{Var}[\xi_A(j,t)] = \frac{1}{N} \mathrm{Var}[\xi_A] = \frac{D_A}{N}.
\end{equation}
Comparing this equation with (\ref{eq:one_neuron_add}), the variance is reduced by the factor of $N=100$. Therefore the dispersion level is reduced from $10^{-2}$ in one neuron to $10^{-4}$ in ESN.

Now we turn on the property of memory making $\gamma=\beta=0.5$. Then sequence of input signal becomes important. To keep the same range of signal and include the property of growing and decreasing input signal, we will use sine function as an input (Fig.~\ref{fig:ESN_uni}e). Figure \ref{fig:ESN_uni}b shows the dispersions for this case. From that moment on we will plot these characteristics depending on the time $t$ to underline the peculiarities of the input signal. Comparing scales of dispersion in panels a and b, one can see that dispersion grows with increasing $\gamma$, as now the noise is accumulated in reservoir. 

The another one thing that should be pointed out is the view of these dependencies. Comparing Fig.~\ref{fig:ESN_uni}a and \ref{fig:ESN_uni}c, there may be an erroneous opinion that now the general form of dependencies has changed a lot. However, it is not so. IF we change the dependence on time in panels (c,d) to the dependence on the mean output value as before, then the general view will be exactly the same as in Fig.~\ref{fig:ESN_uni}a,b.

\begin{figure}[!ht]
\centering\includegraphics[width=1\linewidth]{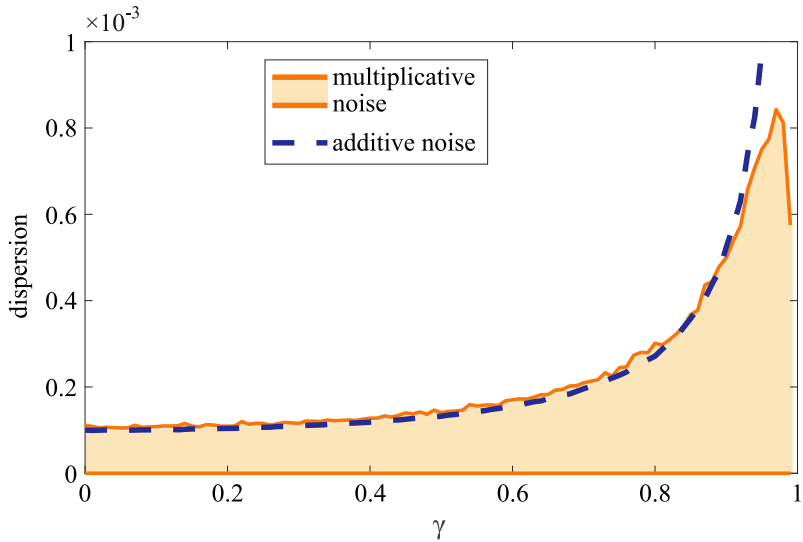} 
\caption{Mean dispersion level (dashed line) for additive noise and dispersion range for multiplicative noise depending on parameter $\gamma$. Other parameters: $\alpha=1$, $\beta=1-\gamma$, $D_A=D_M=10^{-2}$.} \label{fig:ESN_uni_gamma}
\end{figure}

If $\gamma$ grows further up to $0.9$, the maximal dispersion level will be increased to $\approx 5\cdot 10^{-4}$. The general view of dispersion dependencies remains the same as it is shown in Fig.~\ref{fig:ESN_uni}b. In the case of additive noise the dispersion dependency is almost constant, while this dependency for multiplicative noise looks like a doubled sine function and covers a range from zero to some maximal level depending on $\gamma$. In order to reveal the impact of parameter $\gamma$, Fig.~\ref{fig:ESN_uni_gamma} shows how mean dispersion level for additive noise and dispersion range for multiplicative noise changes depending on the parameter $\gamma$.

A uniform reservoir connection matrix $\mathbf{W}^\mathrm{res}$ may look like a rather degenerate case. However, this matrix is sometimes set randomly and does not change during the training process. According to our previous studies \cite{Semenova2022}, in terms of network noise accumulation, a similar uniform connectivity can be interpreted as a matrix with random values and a mean value of $1/N$. Thus, the conclusion ``the smallest level of dispersion in this case can be obtained with weak memory of reservoir and then grows exponentially with parameter $\gamma$ responsible for memory property'' also holds for uniform random $\mathbf{W}^\mathrm{res}$.

\section{ESN with diagonal-like connection matrix}\label{sec:ESN_eye}
\begin{figure}[!ht]
\centering\includegraphics[width=1\linewidth]{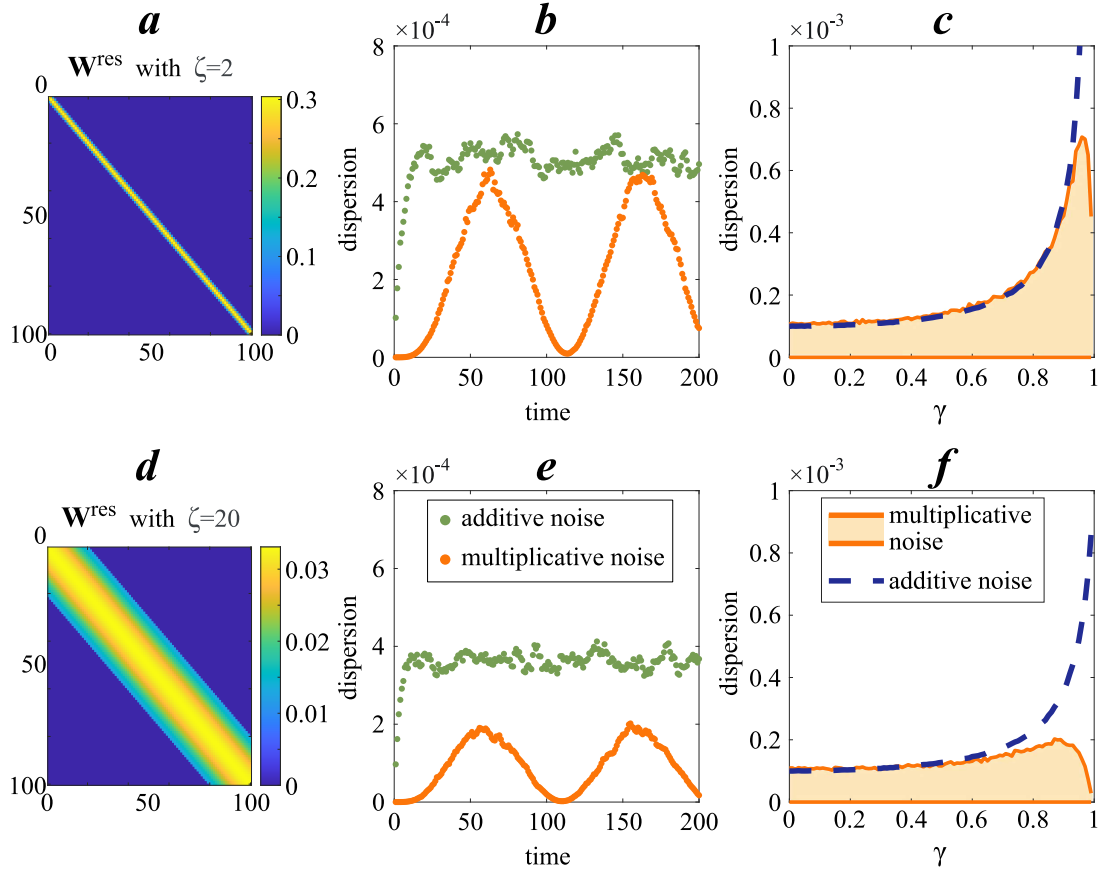} 
\caption{Dispersion of the output signal of ESN with diagonal matrices $\mathbf{W}^\mathrm{res}$. Panels a and d show the connection matrices with $\zeta=2$ and $\zeta=20$, respectively. Panel b shows the dispersion for additive (green) and multiplicative (orange) noise in the case of $\zeta=2$. Panel c shows how this dispersion changes depending on $\gamma$. Panels d,e are the same but for $\zeta=20$. Other fixed parameters: $\alpha=1$, $\gamma=0.8$, $\beta=1-\gamma$, $D_A=D_M=10^{-2}$.} \label{fig:ESN_eye}
\end{figure}

As it has been mentioned in previous section, the reservoir connection matrix $\mathbf{W}^\mathrm{res}$ can be set without change during training process. Usually, it is set to be uniform or diagonal-like \cite{Lukosevicius2009}. In this section we consider the last type of the network. Figure \ref{fig:ESN_eye}a,d shows the diagonal matrices which we will use in reservoir. Both networks are set with some {\it blurring} coefficient $\zeta$. We set this ``blurring'' effect using Gaussian function. For example, in Fig.~\ref{fig:ESN_eye}a this coefficient is set $\zeta=2$ meaning that main diagonal and two terms from the left and right sides of the main diagonal are set according Gaussian function, while the rest are set to be zero. In order to keep the same range of values, we need to make the sum of the elements in each row and column of the matrix $\mathbf{W}^\mathrm{res}$ equal to one, as before. Therefore, the non-zero elements of this matrix are set in the next way:
\begin{equation}
    W^\mathrm{res}_{k,i}=\frac{e^{-(k/\zeta^2)}}{\sum\limits^{\zeta}_{j=-\zeta}e^{-(j/\zeta^2)}}, \ \ \ \ k\in[i-\zeta;i+\zeta].
\end{equation}

In this section we will consider two diagonal-like matrices $\mathbf{W}^\mathrm{res}$ with two ``blurring'' coefficients $\zeta=2$ (Fig.~\ref{fig:ESN_eye}a) and $\zeta=20$ (Fig.~\ref{fig:ESN_eye}d).

Figures \ref{fig:ESN_eye}b,e show the dispersion of the output signal for reservoir connection matrices with $\zeta=2$ and $\zeta=20$ given in corresponding left panels and parameter $\gamma=0.8$. There is a clear difference between dispersion in case of diagonal matrices and dispersion obtained for uniform matrices (Fig.~\ref{fig:ESN_uni}b). The form of $\sigma^2$-dependencies for additive and multiplicative noise remains almost the same, but there is a clear quantitative difference. In the case of uniform connection matrix and diagonal matrix with small $\zeta$ (Fig.~\ref{fig:ESN_eye}b) the maximal value of dispersion for multiplicative noise coincided with its mean value for additive noise. These dependencies are moving away form each other faster for diagonal matrix (Fig.~\ref{fig:ESN_eye}c) comparing with uniform matrix (Fig.~\ref{fig:ESN_uni_gamma}).

These dependencies drift apart faster with larger ``blurring'' coefficient $\zeta=20$ (see Figs.~\ref{fig:ESN_eye}e,f).

Comparing the impact of additive noise in Figs.~\ref{fig:ESN_uni_gamma} and \ref{fig:ESN_eye}c, one can see that dispersions for additive and multiplicative noise depending on $\gamma$ are very similar for diagonal matrix with small ``blurring'' coefficient (Fig.~\ref{fig:ESN_eye}c) and uniform connection matrix (Fig.~\ref{fig:ESN_uni_gamma}). Moreover, the noise is accumulated less for large ``blurring'' coefficient $\zeta$ (Fig.~\ref{fig:ESN_eye}f). In the case of large $\gamma$ the final dispersion level becomes less with growing $\zeta$. This difference is more clear for multiplicative noise. Comparing dispersion ranges in panels c and f of Fig.~\ref{fig:ESN_eye}, one can see that multiplicative dispersion level for large $\zeta=20$ is much less than for small $\zeta=2$. 

\section*{Conclusion and discussion}
In this paper we have studied the impact of uncorrelated additive and multiplicative noise on echo state neural network. The noise was added only to neurons inside the reservoir. These neurons had linear activation function. To analyse the output level of noise we mainly used a dispersion and a signal-to-noise ratio derived from it. 

The parameter $\alpha$ controlling the slope of activation function has no impact on accumulation of additive noise. At the same time, the dispersion for multiplicative noise has a quadratic dependency on $\alpha$ and input signal. The dispersion can be predicted analytically by Eqs.~\ref{eq:one_neuron_add}--\ref{eq:one_neuron_mul}. In our next studies we plan to consider other types of activation functions such as sigmoid functions and piecewise linear activation function.

ESNs are usually set with random uniform reservoir connection matrix or diagonal-like matrix, which are not changed during training process. Therefore, in this paper we considered both types of connection matrices and study the impact of memory on accumulation of noise. We have found some interesting results for these matrices. The noise is less accumulated in ESN with diagonal reservoir connection matrix $\mathbf{W}^\mathrm{res}$ with large ``blurring'' coefficient. Especially it concerns uncorrelated multiplicative noise. The accumulation of noise in uniform $\mathbf{W}^\mathrm{res}$ is almost the same as in diagonal $\mathbf{W}^\mathrm{res}$ with small ``blurring'' coefficient. 

%\bibliographystyle{siam}
%\bibliography{my_ref}

\begin{thebibliography}{23}\itemsep=2pt
\bibitem{LeCun2015} LeCun, Y., Bengio, Y., and Hinton, G., Deep learning, \textit{Nature}, 2015, vol.\,521, no.\,7553, pp.\,436--444.
\bibitem{Maturana2015} Maturana, D., and Scherer, S., VoxNet: A 3D Convolutional Neural Network for real-time object recognition, in \textit{2015 IEEE/RSJ International Conference on
  Intelligent Robots and Systems (IROS)}, IEEE, 2015, pp.\,922--928.
\bibitem{Krizhevsky2017} Krizhevsky, A., Sutskever, I., and Hinton, G.\,E., ImageNet Classification with Deep Convolutional Neural Networks, \textit{Commun. ACM}, 2017, vol.\,60, no.\,6, pp.\,84--90 .
\bibitem{Graves2013} Graves, A., Mohamed, A., and Hinton, G., Speech recognition with deep recurrent neural networks, in \textit{2013 IEEE International Conference on Acoustics, Speech and Signal Processing}, IEEE, 2013, pp.\,6645--6649
\bibitem{Kar2009} Kar, S., and Moura, J.\,M.\,F., Distributed Consensus Algorithms in Sensor Networks With Imperfect Communication: Link Failures and Channel Noise, \textit{IEEE Transactions on Signal Processing}, 2009, vol.\,57, no.\,1, pp.\,355--369.
\bibitem{Brunner2013} Brunner, D., Soriano, M.\,C., Mirasso, C.\,R., and Fischer, I., Parallel photonic information processing at gigabyte per second data rates using transient states, \textit{Nature communications}, 2013, vol.\,4, p.\,1364.
\bibitem{Tuma2016} Tuma, T., Pantazi, A., Le Gallo, M., Sebastian, A., and Eleftheriou, E., Stochastic phase-change neurons, \textit{Nature Nanotechnology}, 2016, vol.\,11, pp.\,693--699.
\bibitem{Torrejon2017} Torrejon, J., Riou, M., Araujo, F.\,A., Tsunegi, S., Khalsa, G., Querlioz, D., Bortolotti, P., Cros, V., Yakushiji, K., Fukushima, A., Kubota, H., Yuasa, S., Stiles, M.\,D. and Grollier, J., Neuromorphic computing with nanoscale spintronic oscillators, \textit{Nature}, 2017, vol.\,547, no.\,7664, pp.\,428-431.
\bibitem{Psaltis1990} Psaltis, D., Brady, D., Gu, X.-G., Lin, S. 
Holography in artificial neural networks, \textit{Nature}, 1990, vol.\,343, no.\,6256, pp.\,325-330 
\bibitem{Bueno2018} Bueno, J., Maktoobi, S., Froehly, L., Fischer, I., Jacquot, M., Larger, L., Brunner, D., Reinforcement Learning in a large scale photonic Recurrent Neural Network, \textit{Optica}, 2018, vol.\,5, no.\,6, pp.\,756 - 760.
\bibitem{Lin2018} Lin, X., Rivenson, Y., Yardimci, N.\,T., Veli, M., Jarrahi, M., Ozcan, A., All-Optical Machine Learning Using Diffractive Deep Neural Networks, \textit{Science}, 2018, vol.\,26, pp.\,1-20.
\bibitem{Shen2017} Shen, Y., Harris, N.\,C., Skirlo, S., Prabhu, M., Baehr-Jones, T., Hochberg, M., Sun, X., Zhao, S., Larochelle, H., Englund, D., and Soljacic, M., Deep Learning with Coherent Nanophotonic Circuits, \textit{Nature Photonics}, 2017, vol.\,11, pp.\,441-446.
\bibitem{Tait2017} Tait, A.\,N., De Lima, T.\,F., Zhou, E., Wu, A.\,X., Nahmias, M.\,A., Shastri, B.\,J., and Prucnal, P.\,R. 
Neuromorphic photonic networks using silicon photonic weight banks, \textit{Scientific Reports}, 2017, vol.\,7, no.\,1, pp.\,1--10.
\bibitem{Moughames2020} Moughames, J., Porte, X., Thiel, M., Ulliac, G., Larger, L., Jacquot, M., Kadic, M., and Brunner, D., Three-dimensional waveguide interconnects for scalable integration of photonic neural networks, \textit{Optica}, 2020, vol.\,7, no.\,6, pp.\,640--646.
\bibitem{Dinc2020} Dinc, N.\,U., Psaltis, D., and Brunner, D., Optical neural networks: The 3D connection \textit{Photoniques}, 2020, no.\,104, pp.\,34--38
\bibitem{Moughames2020a} Moughames, J., Porte, X., Larger, L., Jacquot, M., Kadic, M., and Brunner, D., 3D printed multimode-splitters for photonic interconnects, \textit{Opt. Mater. Express}, 2020, vol.\,10, no.\,11, pp.\,2952--2961.
\bibitem{MourgiasAlexandris2022}
Mourgias-Alexandris, G., Moralis-Pegios, M., Tsakyridis, A., Simos, S., Dabos, G., Totovic, A., Passalis, N., Kirtas, M., Rutirawut, T., Gardes, F.\,Y., Tefas, A.,  Pleros, N., Noise-resilient and high-speed deep learning with coherent silicon photonics \textit{Nature Communications}, 2022, vol.\,13, no.\,1, pp.\,5572. 
\bibitem{Wang2022} Wang, T., Ma, S.-Y., Wright, L.\,G., Onodera, T., Richard, B.\,C., and McMahon, P.\,L., An optical neural network using less than 1 photon per multiplication, \textit{Nature Communications}, 2022, vol.\,13, no.\,1, pp.\,123.
\bibitem{Semenova2019} Semenova, N., Porte, X., Andreoli, L., Jacquot, M., Larger, L., and Brunner, D., Fundamental aspects of noise in analog-hardware neural networks \textit{Chaos}, 2019, vol.\,29, no.\,10, pp.\,103128.
\bibitem{Semenova2022} Semenova, N., Larger, L., and Brunner, D., Understanding and mitigating noise in trained deep neural networks \textit{Neural Networks}, 2022, vol.\,146, pp.\,151--160.
\bibitem{Semenova2022a} Semenova, N., and Brunner, D., Noise-mitigation strategies in physical feedforward neural networks \textit{Chaos}, 2022, vol.\,32, no.\,6, pp.\,061106.
\bibitem{snr} Johnson, D.\,H., Signal-to-noise ratio \textit{Scholarpedia}, 2006, vol.\,1, no.\,12, p\,2088.
\bibitem{Lukosevicius2009} Luko{\v s}evi{\v c}ius, M., and Jaeger, H., Reservoir computing approaches to recurrent neural network training \textit{Computer Science Review}, 2009, vol.\,3, no.\,3, pp.\,127--149.
\end{thebibliography}
\section*{Acknowledgements}
This work was supported by the Russian Science Foundation (Project no. 21-72-00002).

\end{document}